\documentclass{egpubl}

\usepackage[T1]{fontenc}
\usepackage{dfadobe}

\usepackage{cite}
\BibtexOrBiblatex
\electronicVersion
\PrintedOrElectronic
\ifpdf \usepackage[pdftex]{graphicx} \pdfcompresslevel=9
\else \usepackage[dvips]{graphicx} \fi

\usepackage{egweblnk}

\makeatletter
\hypersetup{pdfsubject={Preprint},pdfkeywords={}}
\def\p@Wileypagenumber{\thepage}
\def\noWileypagenumber{\def\p@Wileypagenumber{\mbox{}}}
\def\ps@titlepage{\let\@mkboth\@gobbletwo
  \def\@oddhead{\small\itshape Preprint\hfill}%
  \def\@oddfoot{}%
  \let\@evenhead\@oddhead
  \let\@evenfoot\@oddfoot
  \let\sectionmark\EmptySectionmark
  \let\subsectionmark\EmptySubsectionmark}
\def\ps@headings{\let\@mkboth\@gobbletwo
  \def\@oddhead{\small \hfill \itshape \@shortauthor\ / \@shorttitle
                \hfill \upshape \textrm{\p@Wileypagenumber}}%
  \def\@evenhead{\small \upshape \textrm{\p@Wileypagenumber} \hfill
                 \itshape \@shortauthor\ / \@shorttitle \hfill}%
  \def\@oddfoot{}%
  \def\@evenfoot{}%
  \def\sectionmark##1{}%
  \def\subsectionmark##1{}}
\ps@headings
\makeatother

\usepackage{amssymb}
\usepackage{amsmath}
\usepackage{booktabs}
\usepackage{multirow}

\title[UniBRep: Learning Unified Geometry and Topology for Image-conditioned B-Rep Generation]%
      {UniBRep: Learning Unified Geometry and Topology\\for Image-conditioned B-Rep Generation}

\author[H. Ying et al.]
{
Haiyang Ying$^1$\orcid{0009-0009-8096-4726},
Allen Tu$^1$\orcid{0009-0002-6532-2863},
Jiaye Wu$^2$\orcid{0009-0006-6634-6575},
Tom Goldstein$^1$\orcid{0000-0003-1660-9307},
and Matthias Zwicker$^1$\orcid{0000-0001-8630-5515}
\\
{\parbox{\textwidth}{\centering
$^1$University of Maryland, College Park, Maryland, USA\qquad
$^2$Amazon, Bellevue, Washington, USA\\[1mm]
}}
\vspace{-7mm}
}

\begin{document}

\teaser{
 \includegraphics[width=1.0\linewidth]{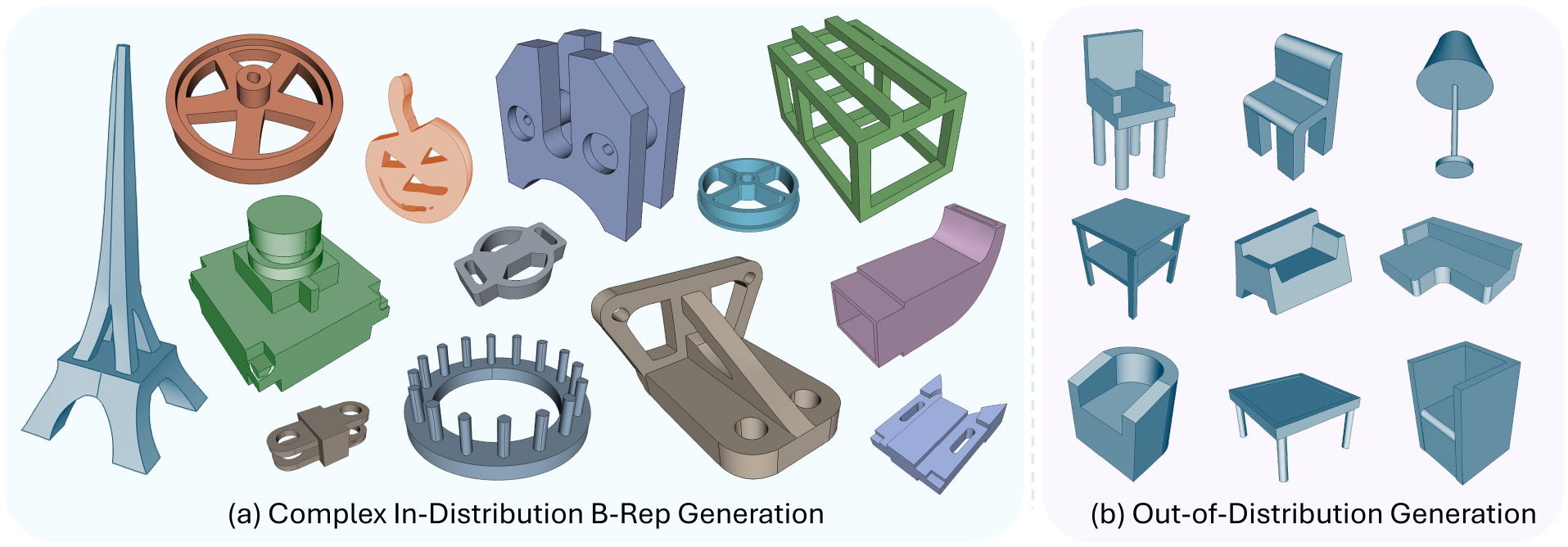}
 \centering
  \caption{
  \textbf{Single-image B-rep generation across shape complexity and distribution shift.}
    UniBRep adapts a pretrained image-to-3D prior to generate feature meshes and recover explicit B-reps without a fixed architectural limit on the number of faces.
    The examples demonstrate (a) generation of complex CAD shapes with detailed face layouts and (b) generalization to objects outside the CAD training distribution.
    }
    \label{fig:teaser}
}

\maketitle
\begin{abstract}
Generating a boundary representation (B-rep) conditioned on a single image requires faithful reconstruction of geometry, valid topology, and support for complex shapes.
We present UniBRep, a geometry-first framework that adapts a pretrained image-to-3D model to generate a feature mesh as a unified intermediate representation.
Its surface provides a geometric scaffold, while spatially aligned learned features encode face-separation cues for topology recovery.
Dual decoder branches generate the geometry and face-separation features; a geometry- and feature-guided construction pipeline then fits parametric surfaces, recovers boundary curves and connectivity, and assembles an explicit B-rep using a CAD kernel.
Recovering topology from mesh regions avoids predefined architectural face-count limits, allowing face count to scale with shape complexity.
On the standard DeepCAD benchmark, UniBRep produces valid B-reps for 80.49\% of inputs and reduces face Chamfer distance from 0.1096 to 0.0345 relative to CADDreamer.
In a matched comparison, UniBRep also outperforms the HoLa public demo across all reported metrics.
Further evaluations demonstrate scalability to high-complexity shapes beyond the standard 30-face range, generalization to objects outside the CAD training distribution, and qualitative transfer to real photographs.



\end{abstract}
\section{Introduction}
\label{sec:intro}

Boundary representations (B-reps) are widely used for 3D modeling in computer-aided design (CAD), with applications spanning manufacturing~\cite{fitter2014review, ji1997machine}, architecture~\cite{chang2013bim, livingston2008cad}, product design~\cite{chang2014product, tovey1989drawing}, and digital content creation.
Unlike implicit fields~\cite{mildenhall2022nerf,wang2021neus} or polygonal meshes, B-reps encode object boundaries using parametric surfaces and boundary curves, together with topology specifying the connectivity among faces, edges, and vertices.
This explicit representation enables precise editing, simulation, and fabrication.
Single-image B-rep generation has recently gained attention in the 3D vision and graphics community.
While conventional 3D generation primarily targets visually plausible geometry, B-rep generation must also produce structurally valid CAD models suitable for downstream use.
Reliable single-image B-rep generation could support reverse engineering, design retrieval, rapid prototyping, and image-based modeling, although recovering complete geometry and topology from a single image remains highly challenging.

This difficulty arises from three coupled requirements: faithful geometry, valid topology, and support for shapes of varying complexity.
Direct B-rep generation methods~\cite{xu2024brepgen,lee2025brepdiff,liu2025hola} jointly predict faces, edges, vertices, and their topological relationships.
Local errors in these predictions can propagate through B-rep assembly, causing gaps, inconsistent adjacency, non-watertight surfaces, or invalid CAD models.
Many such methods also generate fixed-size or bounded sets of B-rep entities, imposing predefined face-count limits that restrict scalability.

Geometry-first methods mitigate these limitations by generating surface geometry before recovering B-rep structure.
CADDreamer~\cite{li2025caddreamer} follows this approach by predicting multi-view normal and primitive maps from a single image, reconstructing a mesh, and converting it into a B-rep through a multi-stage post-processing pipeline.
Although this formulation avoids a predefined face-count limit, its final quality depends on both the fidelity of the generated geometry and the robustness of the subsequent B-rep construction, where errors may accumulate across stages.
Moreover, geometry generators trained only on CAD data may lack the broad 3D priors needed to generalize across diverse inputs.

We propose \textbf{UniBRep}, a geometry-first framework for single-image-conditioned B-rep generation.
UniBRep adapts TRELLIS~\cite{xiang2024structured}, a pretrained image-to-3D foundation model, to generate a feature mesh as a unified intermediate representation: its surface provides a geometric scaffold, while its learned features encode face-separation cues for topology recovery.
A reconstruction pipeline converts this representation into an explicit B-rep by fitting parametric surfaces and recovering boundary curves and connectivity.
By recovering topology from generated mesh regions, UniBRep largely preserves the generated geometry while supporting a variable number of faces without a predefined architectural limit.
Experiments demonstrate improved geometric and topological quality, B-rep validity, and geometry preservation over prior methods.
Further evaluations show that UniBRep scales to high-complexity shapes, generalizes to objects outside the CAD training distribution, and can be applied to real-world images.

In summary, our contributions are:
\begin{itemize}
    \item A geometry-first framework that adapts a pretrained image-to-3D prior to generate feature meshes as a unified representation for B-rep geometry and topology recovery.
    \item A B-rep reconstruction pipeline that converts generated feature meshes into high-quality CAD models while allowing face count to scale with shape complexity.
    \item Strong B-rep generation performance, scalability to high-complexity shapes, and generalization to out-of-distribution objects and real-world images.
\end{itemize}
\section{Related Work}
\label{sec:related}

\begin{figure*}[t]
    \centering
    \includegraphics[width=\linewidth]{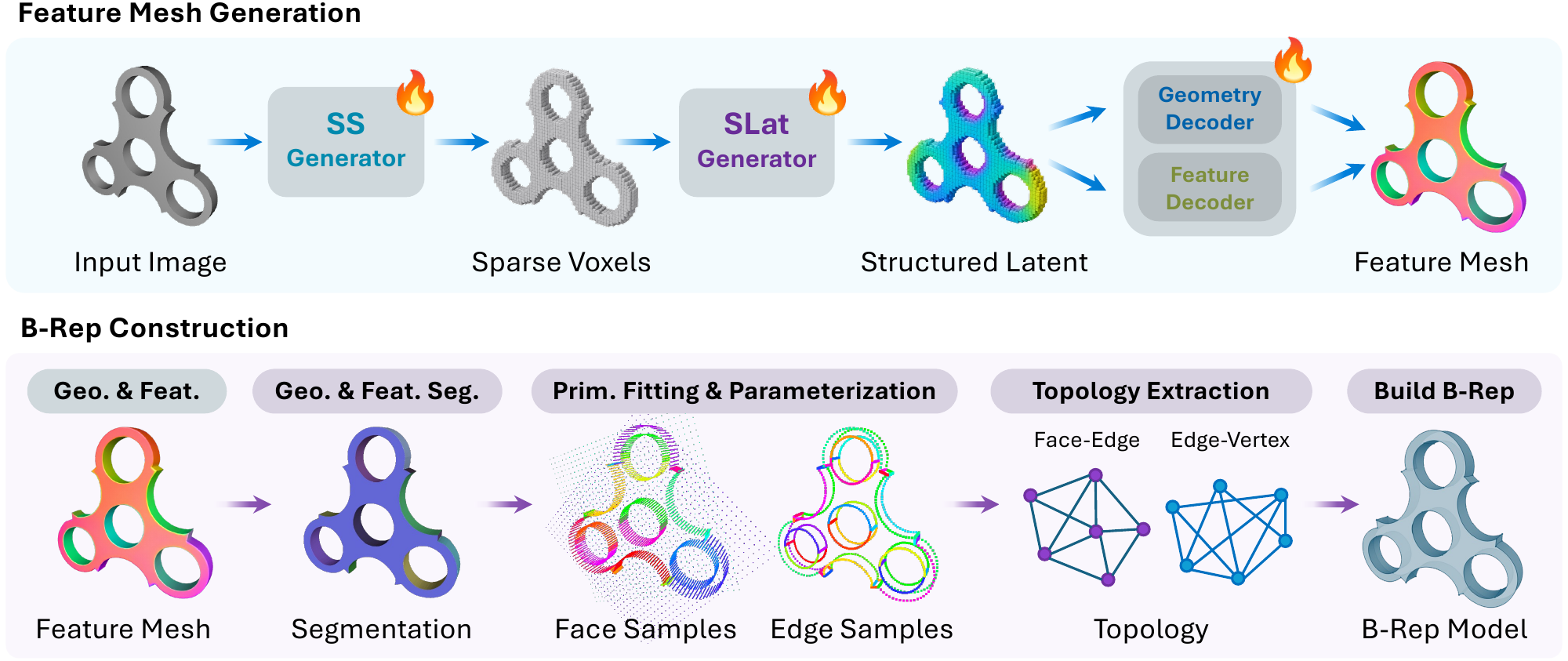}
    \caption{\textbf{Overview of UniBRep.}
    Top: UniBRep adapts TRELLIS's sparse-structure (SS) and structured latent (SLat) generators to produce a CAD-adapted SLat from a single input image.
    Dual geometry and feature SLat decoder branches map the shared latent to surface geometry and learned face-separation features, forming a feature mesh.
    Bottom: The construction pipeline partitions the feature mesh using geometry- and feature-guided segmentation, fits and parameterizes surfaces, extracts topology, and assembles an explicit B-rep using a CAD kernel.}
    \label{fig:method_overview}
\end{figure*}

Recent CAD generation methods broadly fall into three categories: construction-program generation, direct B-rep generation, and geometry generation followed by CAD construction.
Program-based approaches represent CAD models as sequences of parametric operations, such as sketching, extrusion, and Boolean operations.
DeepCAD~\cite{Wu_2021_ICCV} introduces a transformer-based framework for CAD program generation, while subsequent work improves controllability and extends this paradigm to image-conditioned generation~\cite{xu2022skexgen,chen2025cadcrafter}.
Although construction programs are interpretable and editable, these methods typically require sequence annotations and predefined operation vocabularies.
These constraints limit their ability to represent shapes involving unsupported operations or surface types, such as fillets, sweeps, cones, spheres, and free-form surfaces.

The second category directly models B-rep geometry and topology through tokens, sequences, graphs, complexes, or latent representations.
HoLa~\cite{liu2025hola} uses a holistic surface latent for conditional B-rep generation on the filtered DeepCAD benchmark.
Other methods generate B-rep surfaces and topological entities using chain complexes, diffusion models, or autoregressive models~\cite{GuoComplexGen2022,xu2024brepgen,lee2025brepdiff,li2025dtgbrepgen,guo2025brepgiff,xu2025autobrep,li2025brepgpt,jayaraman2022solidgen}.
However, directly generating structured B-rep entities or fixed-size surface representations can make assembly sensitive to local errors, leading to gaps, inconsistent connectivity, or invalid B-reps.
Several methods also impose predefined limits on the number of faces or topological entities, restricting their ability to scale with shape complexity.
UniBRep instead recovers topology from generated mesh regions, allowing face count to scale with shape complexity without a predefined architectural limit.

Finally, geometry-first methods construct structured CAD representations from intermediate geometry.
CADDreamer~\cite{li2025caddreamer} predicts multi-view normal and primitive maps from a single image, generates a mesh, and constructs a B-rep from optimized analytic primitive intersections.
Point2CAD~\cite{liu2024point2cad} and NVDNet~\cite{liu2024nvdnet} recover parametric primitives from point clouds.
Although these methods avoid direct B-rep entity generation, errors in the geometric scaffold can propagate through subsequent B-rep construction.
UniBRep instead adapts TRELLIS~\cite{xiang2024structured}, whose large-scale image-to-3D pretraining provides a broad geometry prior that supports generalization beyond the CAD training distribution.
It uses TRELLIS's structured latent representation and FlexiCubes-based decoding~\cite{shen2023flexicubes} to generate feature meshes that combine surface geometry with learned face-separation cues.
A B-rep construction pipeline then fits analytic and free-form surfaces and recovers topology from the feature mesh while largely preserving the generated geometry.

\section{Method}
\label{sec:method}

Given a single image, UniBRep first generates a feature mesh using a CAD-adapted image-to-3D foundation model.
This unified representation combines a surface mesh as a geometric scaffold with learned face-separation features for topology recovery.
UniBRep then constructs an explicit B-rep through geometry- and feature-guided segmentation, analytic and free-form surface fitting, boundary and connectivity recovery, and CAD-kernel assembly.
Fig.~\ref{fig:method_overview} illustrates the UniBRep pipeline.

\subsection{Feature Mesh Representation}
\label{subsec:feature_mesh}

UniBRep represents the intermediate output of image-conditioned generation as a feature mesh $\mathcal{G}=(\mathcal{M},\mathbf{F})$, where \(\mathcal{M}=(V_{\mathcal{M}},E_{\mathcal{M}},T_{\mathcal{M}})\) is a triangular surface mesh and \(\mathbf{F}\) is a feature field aligned with its surface. For each surface point \(\mathbf{x}\in\mathcal{M}\), the feature \(\mathbf{f}(\mathbf{x})\in\mathbb{R}^{d_f}\) is a learned descriptor of its B-rep face membership. The mesh captures the generated geometry, while the feature field provides cues for recovering its face partition.

The feature field is learned as an instance embedding rather than a fixed semantic labeling. Features associated with points on the same B-rep face are encouraged to be similar, whereas features belonging to different faces are separated. Consequently, the surface features can be clustered into a variable number of candidate face regions without requiring predefined face labels or an upper bound on the number of faces.

The feature mesh does not directly encode the complete B-rep topology.
Instead, it provides a shared geometric domain that spatially aligns face-partition cues with the generated surface.
Boundary curves, vertices, and their connectivity are subsequently recovered from the partitioned mesh.

\subsection{Image-to-Feature-Mesh Generation}
\label{subsec:feature_mesh_generation}

We build on TRELLIS~\cite{xiang2024structured}, a pretrained image-to-3D foundation model based on Structured Latent Representations (SLat). Given an input image \(I\), the sparse-structure (SS) generator $G_{\mathrm{SS}}$ first predicts a set of active voxels \(\mathcal{V}\), then the SLat generator $G_{\mathrm{SLat}}$ predicts a latent feature \(\mathbf{z}_i\in\mathbb{R}^{d}\) for each active voxel \(\mathbf{p}_i\):
\begin{equation}
\mathcal{V}=G_{\mathrm{SS}}(I),
\qquad
\mathbf{z}_{\mathcal{V}}
=
G_{\mathrm{SLat}}(I,\mathcal{V})
=
\{(\mathbf{p}_i,\mathbf{z}_i)\}_{\mathbf{p}_i\in\mathcal{V}}.
\end{equation}
The active voxels describe the coarse spatial structure of the object, while their latent features encode local geometric information.
The SS and SLat generators use image-conditioned transformer backbones, while the SLat decoder $D_{\mathrm{SLat}}$ progressively upsamples the sparse features into dense output fields.

We adapt the SS and SLat generators to the CAD domain by inserting low-rank adaptation (LoRA) modules~\cite{hu2022lora} into their transformer blocks.
Following TRELLIS's training formulation, we optimize both generators using conditional flow-matching objectives on structured latent targets constructed from tessellated CAD models.
TRELLIS's large-scale pretraining on 3D assets provides a broad geometric prior that supports generalization beyond the CAD shapes used for adaptation.

\subsubsection{Dual-Branch SLat Decoding}
The original TRELLIS SLat decoder $D_{\mathrm{SLat}}$ maps the generated SLat $\mathbf{z}_{\mathcal{V}}$ to geometry and appearance attributes.
We adapt it into separate geometry and feature branches to decouple geometry generation from face-feature learning:
\begin{equation}
\boldsymbol{\gamma}
=
D_g(\mathbf{z}_{\mathcal{V}}),
\qquad
\mathbf{H}
=
D_f(\mathbf{z}_{\mathcal{V}}),
\end{equation}
where $D_g$ and $D_f$ are the geometry and feature decoders, respectively. Here, \(\boldsymbol{\gamma}\) contains the geometric quantities required by FlexiCubes~\cite{shen2023flexicubes}, while \(\mathbf{H}\) is a dense grid of eight-dimensional face-separation features. Both branches are initialized from the same pretrained TRELLIS decoder but use independent LoRA parameters and output layers.

The geometry branch predicts signed-distance values, deformation offsets, and blending weights at \(256^3\) resolution, from which FlexiCubes extracts the surface mesh $\mathcal{M}=\operatorname{FC}(\boldsymbol{\gamma})$.
The dense grid \(\mathbf{H}\) defines a spatial feature field through interpolation. After extracting \(\mathcal{M}\), we evaluate this field at its vertices using the same interpolation procedure employed by TRELLIS for vertex colors:
\begin{equation}
\mathbf{f}_i
=
\operatorname{Interp}(\mathbf{H},\mathbf{v}_i),
\qquad
\mathbf{v}_i\in V_{\mathcal{M}}.
\end{equation}
The resulting features are spatially aligned with the generated surface and used by the downstream B-rep construction pipeline.

We train the decoder branches sequentially.
The geometry branch is first adapted to CAD geometry and then frozen before the feature branch is trained.
This prevents face-feature supervision from modifying the generated surface and allows the distinct objectives of geometry generation and face-feature discrimination to be optimized separately.

\subsubsection{Geometry and Face-Feature Supervision}
We train UniBRep using B-rep models from DeepCAD~\cite{Wu_2021_ICCV}.
Each source B-rep is tessellated while preserving the correspondence between mesh triangles and their source B-rep faces. The geometry branch follows TRELLIS's mesh-decoder objective using the tessellated CAD geometry and its rendered views.

For feature supervision, we densely sample the ground-truth mesh, assigning each sample the identifier of its source B-rep face.
We transfer these labels to samples from the generated surface through nearest-neighbor matching.
Let \(\{\mathcal{F}_k\}_{k=1}^{K}\) denote the resulting partition of supervised samples into \(K\) B-rep faces.
For each face, we compute its feature center:
\begin{equation}
\bar{\mathbf{f}}_k
=
\frac{1}{|\mathcal{F}_k|}
\sum_{\mathbf{x}_i\in\mathcal{F}_k}
\mathbf{f}(\mathbf{x}_i).
\end{equation}
Following HPNet~\cite{yan2021hpnet} and SED-Net~\cite{li2023sednet}, the pull term encourages features within each face to remain compact:
\begin{equation}
\mathcal{L}_{\mathrm{pull}}
=
\frac{1}{K}
\sum_{k=1}^{K}
\frac{1}{|\mathcal{F}_k|}
\sum_{\mathbf{x}_i\in\mathcal{F}_k}
\max\left(
\|\mathbf{f}(\mathbf{x}_i)-\bar{\mathbf{f}}_k\|-\delta_v,0
\right),
\end{equation}
while the push term separates the centers of different faces:
\begin{equation}
\mathcal{L}_{\mathrm{push}}
=
\frac{2}{K(K-1)}
\sum_{k<l}
\max\left(
\delta_d-\|\bar{\mathbf{f}}_k-\bar{\mathbf{f}}_l\|,0
\right).
\end{equation}
The complete feature objective is
\begin{equation}
\mathcal{L}_{\mathrm{feat}}
=
\lambda_{\mathrm{pull}}\mathcal{L}_{\mathrm{pull}}
+
\lambda_{\mathrm{push}}\mathcal{L}_{\mathrm{push}}.
\end{equation}
This objective learns instance embeddings for B-rep faces rather than fixed semantic labels.
These embeddings can then be clustered into a variable number of face regions, allowing the recovered face count to scale with shape complexity.

\subsection{B-Rep Construction from Feature Meshes}
\label{subsec:brep_construction}

Given the generated feature mesh \(\mathcal{G}=(\mathcal{M},\mathbf{F})\), we construct an explicit B-rep through geometry-guided segmentation, feature-guided refinement, surface fitting, boundary and topology recovery, and CAD-kernel assembly.

\begin{figure}[t]
    \centering
    \includegraphics[width=\linewidth]{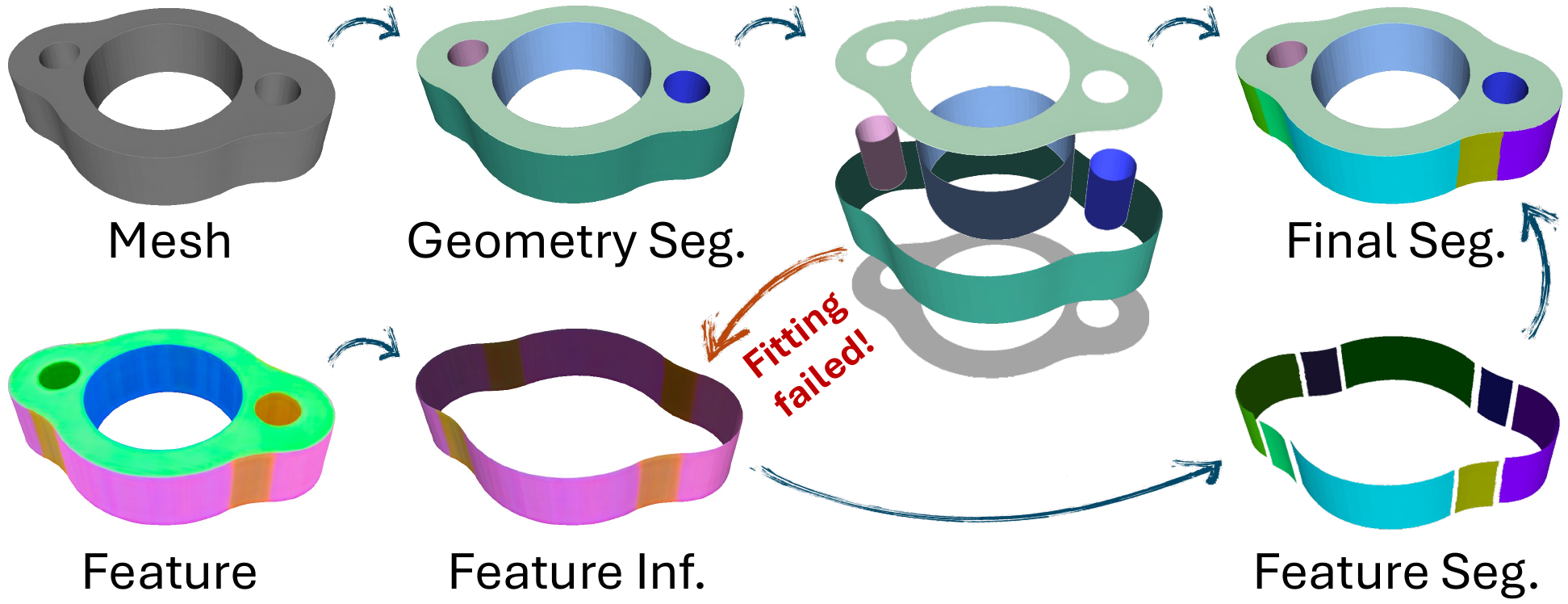}
    \caption{
    \textbf{Geometry- and feature-guided segmentation.}
    We first partition the generated mesh using local geometric cues and fit analytic primitives to the resulting regions.
    Regions that fail primitive fitting are further segmented using learned face-separation features and then refitted.
    }
    \label{fig:segmentation_refinement}
\end{figure}

\subsubsection{Geometry-Guided Segmentation and Surface Fitting}
We first partition \(\mathcal{M}\) into connected candidate face regions using region growing~\cite{adams1994regiongrowing} based on mesh adjacency, boundary strength, and normal consistency.
Because this procedure operates directly on the mesh, it produces a variable number of regions without assuming a predefined face count.

For each region, we sequentially fit planes, cylinders, cones, spheres, and tori using RANSAC~\cite{schnabel2007ransac}.
A primitive is accepted only if it meets the prescribed thresholds for support ratio and geometric residual.
Regions that cannot be adequately represented by any supported primitive are passed to feature-guided refinement.

\subsubsection{Feature-Guided Refinement and Free-Form Fitting}
For each region that fails initial primitive fitting, we cluster its learned per-vertex features using mean shift.
Because the feature embedding is trained to group samples from the same B-rep face and separate samples from different faces, the resulting clusters reveal subdivisions that may not be apparent from local geometry.

A feature cluster may contain spatially disconnected parts of the mesh.
We therefore divide each cluster into connected components based on mesh adjacency, treat each component as a candidate region, and repeat analytic primitive fitting using the same criteria.
Regions successfully fitted during the initial geometry-based stage remain unchanged, so feature guidance is introduced only where geometric fitting is insufficient.

A refined region that still cannot be represented by a supported analytic primitive is parameterized using FlattenAnything~\cite{zhang2024flatten} and fitted with a B-spline surface.
This fallback enables the reconstruction of irregular and free-form regions without forcing them into the predefined analytic surface families.

\begin{table*}[t]
\centering
\normalsize
\caption{
\textbf{Quantitative comparison of single-image-conditioned B-rep generation on DeepCAD~\cite{Wu_2021_ICCV}.}
The $N{=}2{,}424$ rows report results on the standard 7--30-face benchmark, while the $N{=}818$ rows report the matched comparison on a randomly selected subset of test objects.
The best results within each set are shown in \textbf{bold}.
$^\ast$ indicates that CADDreamer's accepted outputs are OpenCascade compounds rather than necessarily watertight single solids.
$^\dagger$ denotes NVDNet evaluated with our generated geometry and Point2CAD with our generated geometry and segmentation; both output parametric surfaces rather than B-reps.
See Sec.~\ref{sec:experiments} for the full evaluation protocol.
}
\label{tab:main_comparison}
\begingroup
\renewcommand{\arraystretch}{1.12}
\begin{tabular}{c|l|ccc|ccc|cc|c}
\toprule
\multirow{2}{*}{$N$} & \multirow{2}{*}{\centering Method}
& \multicolumn{3}{c|}{Chamfer Distance~$\downarrow$}
& \multicolumn{3}{c|}{Geometry F-score~$\uparrow$}
& \multicolumn{2}{c|}{Topology~$\uparrow$}
& \multirow{2}{*}{Valid Ratio~$\uparrow$} \\
& & Vertex & Edge & Face
& Vertex & Edge & Face
& FE & EV
& \\
\midrule

\multirow{4}{*}{$2{,}424$}
& CADDreamer~\cite{li2025caddreamer}
& 0.3135 & 0.1824 & 0.1096
& 0.425 & 0.359 & 0.607
& 0.384 & 0.483
& 18.28\%$^\ast$ \\

& NVDNet$^\dagger$~\cite{liu2024nvdnet}
& 0.2100 & 0.0895 & 0.0458
& 0.572 & 0.662 & 0.747
& 0.694 & 0.620
& -- \\

& Point2CAD$^\dagger$~\cite{liu2024point2cad}
& 0.2382 & 0.1127 & 0.0615
& 0.517 & 0.621 & 0.708
& 0.725 & 0.633
& -- \\

& \textbf{UniBRep (Ours)}
& \textbf{0.1328} & \textbf{0.0627} & \textbf{0.0345}
& \textbf{0.742} & \textbf{0.790} & \textbf{0.844}
& \textbf{0.812} & \textbf{0.783}
& \textbf{80.49\%} \\

\midrule

\multirow{2}{*}{$818$}
& HoLa (demo)~\cite{liu2025hola}
& 0.3227 & 0.1737 & 0.1028
& 0.213 & 0.420 & 0.560
& 0.651 & 0.340
& 70.66\% \\

& \textbf{UniBRep (Ours)}
& \textbf{0.1277} & \textbf{0.0593} & \textbf{0.0327}
& \textbf{0.754} & \textbf{0.802} & \textbf{0.853}
& \textbf{0.816} & \textbf{0.782}
& \textbf{81.30\%} \\

\bottomrule
\end{tabular}
\endgroup
\end{table*}

\subsubsection{Boundary Recovery and B-Rep Assembly}
After assigning a fitted surface to every face region, we recover B-rep boundaries directly from the segmented mesh. For each pair of adjacent regions, we trace their shared mesh boundary to obtain a candidate edge polyline. Unlike pipelines based on explicit surface--surface intersection~\cite{liu2024point2cad}, this procedure preserves the adjacency of the generated mesh and is less sensitive to small discrepancies between independently fitted surfaces.

Following BrepGen~\cite{xu2024brepgen}, we approximate each boundary polyline with a cubic B-spline curve and resample it along its arc length.
We fix endpoints shared by multiple curves to common vertex positions to maintain consistency among neighboring trimming loops.
The resulting incidences among face regions, boundary curves, and shared endpoints determine the face--edge and edge--vertex connectivity.

For every face incident to a recovered 3D boundary curve, we project the curve onto its fitted surface to obtain a parametric curve that defines the trimming boundary.
Using the recovered topology, we assemble the trimmed faces, shared curves, and vertices into a B-rep and then sew and heal it with OpenCascade.

The resulting B-rep explicitly represents faces, edges, vertices, surfaces, curves, and their incidence relations, enabling standard CAD-kernel operations.
Because topology is recovered from a mesh partition rather than generated as a bounded collection of entities, UniBRep does not impose a predefined architectural upper bound on the number of faces.
Moreover, local surface fitting and direct boundary tracing largely preserve the geometry of the generated feature mesh.
We evaluate this preservation in Sec.~\ref{subsec:component_analysis}.
Detailed B-rep construction parameters are provided in Appendix~\ref{app:implementation_details}.
\section{Experiments}
\label{sec:experiments}

\begin{figure*}[t]
    \centering
    \includegraphics[width=\linewidth]{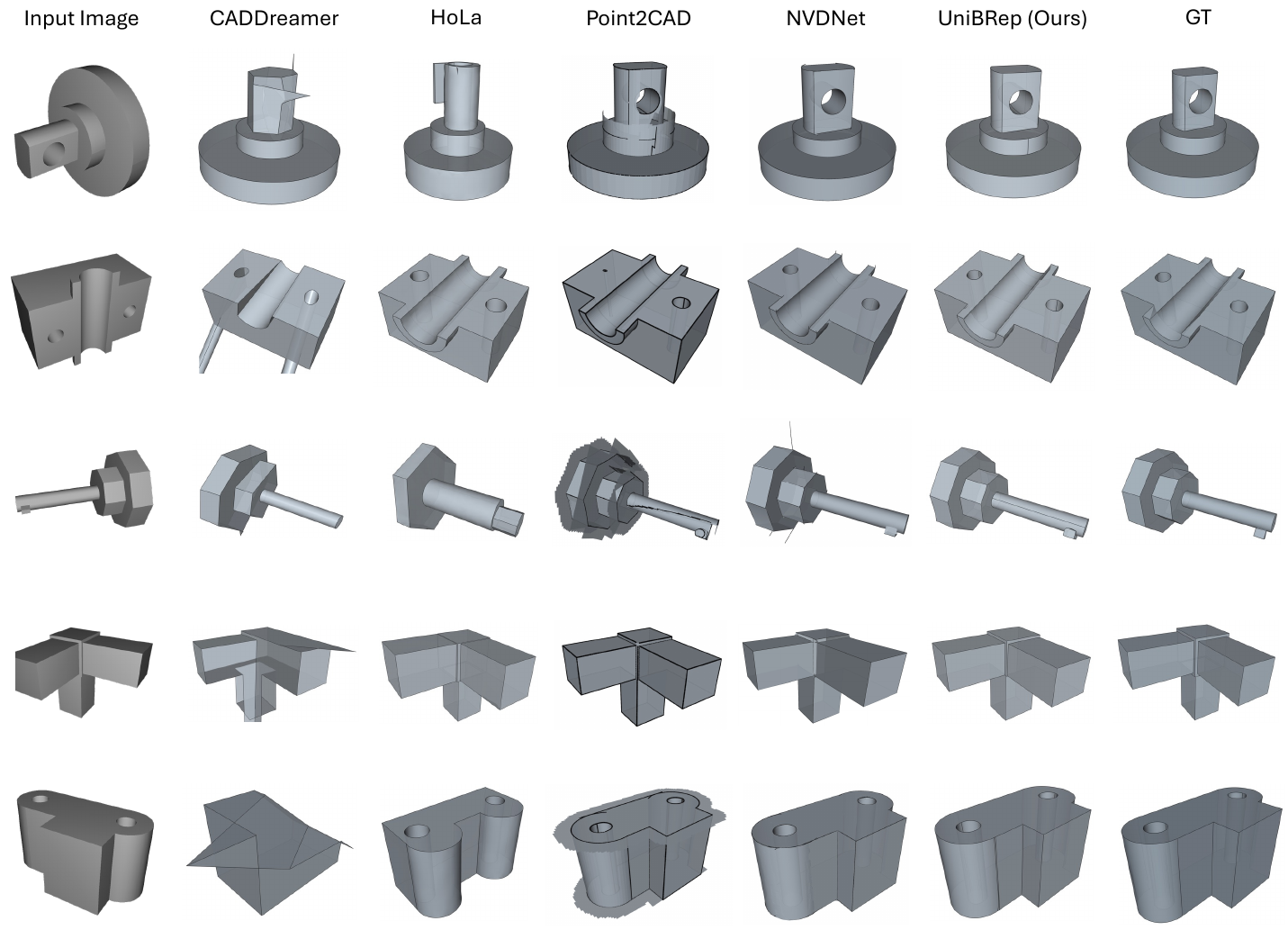}
    \caption{
    \textbf{Qualitative comparison on DeepCAD~\cite{Wu_2021_ICCV}.}
    Outputs and ground truth are rendered from a common evaluation view that differs from the input view.
    Method-specific inputs are described in Sec.~\ref{subsec:baselines}.
    UniBRep more faithfully recovers fine geometry and CAD structure, while the baselines~\cite{li2025caddreamer,liu2025hola,liu2024point2cad,liu2024nvdnet} often omit parts or produce disconnected and non-watertight surfaces.
    }
    \label{fig:qualitative_deepcad_main}
\end{figure*}

We train UniBRep by finetuning TRELLIS~\cite{xiang2024structured} on DeepCAD~\cite{Wu_2021_ICCV}, a filtered subset of ABC~\cite{Koch_2019_CVPR}. We use BRepGen's~\cite{xu2024brepgen} data split and follow HoLa~\cite{liu2025hola} by excluding models with fewer than seven B-rep faces, yielding 64,049 training, 4,466 validation, and 2,605 test models.
Each source B-rep is tessellated with OpenCascade and normalized to $[-0.5,0.5]^3$.
For each training model, we render 150 random views for SLat targets and 25 random views for image conditioning; inference uses a single RGB image.
The SS and SLat generators use rank-32 LoRA modules, while the geometry and feature decoder branches use rank 16.
Appendix~\ref{app:implementation_details} provides additional implementation details.

\subsection{Baselines}
\label{subsec:baselines}
We compare with CADDreamer~\cite{li2025caddreamer}, a geometry-first image-conditioned CAD generation method, and HoLa~\cite{liu2025hola}, a direct conditional B-rep generator.
For each comparison, we use the same underlying test object and viewing direction but render the input required by each pipeline.
UniBRep receives an RGB image, while CADDreamer receives an orthographically rendered ground-truth normal map.
HoLa does not release the model weights, inference code, or evaluation code needed to reproduce its published results, so we evaluate its public Hugging Face demo using an RGB rendering from the same viewing direction with black edge overlays matching the input style shown in its paper.
We evaluate the HoLa demo and UniBRep outputs using the common protocol described in Sec.~\ref{subsec:metrics}; Appendix~\ref{app:evaluation_details} compares the demo results with HoLa's published results and analyzes the protocol discrepancy.
HoLa's released configuration supports at most 30 faces, whereas UniBRep has no predefined architectural face-count cap.

We additionally evaluate point-cloud-based reconstruction methods NVDNet~\cite{liu2024nvdnet} and Point2CAD~\cite{liu2024point2cad} as controlled baselines using UniBRep intermediate outputs.
NVDNet receives points sampled from our generated geometry, while Point2CAD receives our generated geometry and segmentation.
Both output parametric surfaces rather than B-reps, so valid-output ratios are not reported.
These comparisons isolate the contributions of our learned face-separation features and B-rep construction strategy.

\subsection{Datasets}
\label{subsec:datasets}

We adopt the 2,605-model DeepCAD~\cite{Wu_2021_ICCV} test split.
Following HoLa~\cite{liu2025hola}, we use the 2,424 models with 7--30 B-rep faces as the standard \emph{DeepCAD} benchmark.
We separately evaluate the remaining 181 models as \emph{DeepCAD 31+ Faces}.
For the matched HoLa comparison, we randomly select 818 objects and one rendered conditioning viewpoint per object from the standard benchmark.
HoLa and UniBRep are evaluated on all resulting object--viewpoint pairs without filtering by output quality.

To evaluate distribution shift, we use 40 objects from the furniture evaluation set of BrepGen~\cite{xu2024brepgen} that lie outside the DeepCAD training distribution.
We also qualitatively evaluate foreground objects segmented from real photographs.
Unless otherwise specified, all evaluations use one generated output per input.

\subsection{Metrics}
\label{subsec:metrics}

Following NVDNet~\cite{liu2024nvdnet}, we match generated and ground-truth B-rep entities using Hungarian matching and compute Chamfer distance and geometry F-score for vertices, edges, and faces.
We also report topology F-score for face--edge (FE) and edge--vertex (EV) connectivity, valid-output ratio, and mesh-segmentation F-score.
To assess geometry preservation, we compare the Chamfer distances of generated meshes and final B-reps to ground truth.
Following BrepDiff~\cite{lee2025brepdiff}, a valid output must be a readable single OpenCascade solid that passes wire and edge checks and can be tessellated into a nonempty mesh; detailed checks are provided in Appendix~\ref{app:evaluation_details}.
For methods that generate B-reps, geometry and topology metrics are averaged over valid outputs.

To support reproducibility, we will publicly release our code, model weights, data splits, and generated outputs.

\section{Results}
\label{sec:results}

\begin{table*}[t]
\centering
\caption{\textbf{Component ablation on DeepCAD~\cite{Wu_2021_ICCV}.}
\emph{Geo. FT} denotes CAD geometry finetuning, \emph{Feat. Ref.} denotes feature-guided segmentation refinement, and \emph{Supervision} denotes RGB or contrastive feature supervision.
Our full configuration achieves the best results.}
\label{tab:components}
\begin{tabular}{cc c|ccc|ccc|cc|c}
\toprule
\multirow{2}{*}{Geo. FT}
& \multirow{2}{*}{Feat. Ref.}
& \multirow{2}{*}{Supervision}
& \multicolumn{3}{c|}{Chamfer Distance~$\downarrow$}
& \multicolumn{3}{c|}{Geometry F-score~$\uparrow$}
& \multicolumn{2}{c|}{Topology~$\uparrow$}
& \multirow{2}{*}{Valid Ratio~$\uparrow$} \\
& &
& Vertex & Edge & Face
& Vertex & Edge & Face
& FE & EV
& \\
\midrule

 &  & --
& 0.1827 & 0.0895 & 0.0506
& 0.627 & 0.682 & 0.757
& 0.731 & 0.701
& 47.65\% \\

\checkmark &  & --
& 0.1564 & 0.0695 & 0.0382
& 0.695 & 0.739 & 0.799
& 0.775 & 0.741
& 62.50\% \\

\checkmark & \checkmark & RGB
& 0.1381 & 0.0640 & 0.0355
& 0.716 & 0.766 & 0.824
& 0.791 & 0.757
& 74.75\% \\

\midrule

\checkmark & \checkmark & Contrastive
& \textbf{0.1328} & \textbf{0.0627} & \textbf{0.0345}
& \textbf{0.742} & \textbf{0.790} & \textbf{0.844}
& \textbf{0.812} & \textbf{0.783}
& \textbf{80.49\%} \\

\bottomrule
\end{tabular}
\end{table*}

Table~\ref{tab:main_comparison} compares UniBRep with the baselines on DeepCAD~\cite{Wu_2021_ICCV}.
On the standard 2,424-object benchmark, UniBRep reduces face Chamfer distance from $0.1096$ to $0.0345$ and raises the valid-output ratio from $18.28\%$ to $80.49\%$ relative to CADDreamer~\cite{li2025caddreamer}.
It also increases face--edge topology F-score from $0.384$ to $0.812$, showing that the gains extend beyond surface geometry to the recovered B-rep structure.
This validity advantage holds under a stricter criterion: CADDreamer accepts compounds, whereas UniBRep requires a valid single solid.

On the matched subset of 818 test objects, UniBRep outperforms the HoLa public demo~\cite{liu2025hola} on every reported metric, increasing the valid-output ratio from $70.66\%$ to $81.30\%$ and reducing face Chamfer distance from $0.1028$ to $0.0327$.
Both methods use the same object--viewpoint pairs and are evaluated with the common protocol described in Sec.~\ref{subsec:metrics}, making this the closest available comparison with HoLa's released system.
However, HoLa's paper reports an $86.37\%$ valid-output ratio and a face Chamfer distance of $0.0101$, revealing a substantial discrepancy between its published and public-demo results.
Because HoLa does not release its model weights, inference and evaluation code, or complete conditioning-view generation procedure, we cannot determine whether the gap stems from camera selection, rendering, preprocessing, or evaluation.
We compare HoLa's published and public-demo results and examine the protocol discrepancy in Appendix~\ref{app:evaluation_details}.

Fig.~\ref{fig:qualitative_deepcad_main} illustrates how CADDreamer often captures coarse shape but distorts surfaces or loses thin structures, while the HoLa demo frequently omits fine details or alters the object structure.
Point2CAD~\cite{liu2024point2cad} and NVDNet~\cite{liu2024nvdnet} produce outputs with disconnected or hanging surfaces and curves.
In contrast, UniBRep more faithfully recovers holes, protrusions, repeated parts, and sharp CAD boundaries.
Additional qualitative comparisons are provided in Appendix~\ref{app:additional_qualitative_results}.

On an NVIDIA RTX~A5000 GPU, UniBRep averages $3.88$ minutes per object, compared with $22.76$ minutes for CADDreamer.
Feature-mesh generation requires only $0.17$ minutes, while surface parameterization is the dominant stage at $2.07$ minutes.

\subsection{Component and Reconstruction Analysis}
\label{subsec:component_analysis}

Table~\ref{tab:components} evaluates the main design choices in UniBRep.
CAD geometry finetuning and feature-guided refinement improve every reported metric, while contrastive supervision outperforms RGB supervision.
Collectively, these components raise validity from $47.65\%$ to $80.49\%$, with the full configuration performing best across geometry and topology metrics.
Because the geometry branch is frozen during feature training, the improvement from contrastive supervision reflects more discriminative face-separation cues rather than changes to the generated surface.

Given the same generated geometry, UniBRep's feature-guided segmentation achieves a mesh-segmentation F-score of $0.783$, compared with $0.690$ for NVDNet~\cite{liu2024nvdnet}, isolating the contribution of the learned feature field from subsequent fitting and assembly.
The controlled comparisons in Table~\ref{tab:main_comparison} further show that UniBRep achieves lower face Chamfer distance and higher face--edge topology F-score than NVDNet and Point2CAD.
These comparisons demonstrate the benefits of UniBRep's learned face partitioning and mesh-boundary-based construction.

Table~\ref{tab:geometry_primitive_analysis} evaluates geometry preservation and primitive accuracy during B-rep construction.
UniBRep's mean Chamfer distance increases only from $0.0319$ for the generated mesh to $0.0345$ for the final B-rep, whereas CADDreamer's nearly doubles.
UniBRep also achieves higher overall and mean per-type primitive accuracy.
Appendix~\ref{app:reconstruction_sensitivity} shows that geometry and topology metrics remain stable under moderate reconstruction parameter changes, with only marginal gains from repeated post-processing.

\begin{table}[t]
\centering
\caption{
\textbf{B-rep reconstruction quality on DeepCAD~\cite{Wu_2021_ICCV}.}
We compare the construction pipelines of CADDreamer~\cite{li2025caddreamer} and UniBRep on cases whose final outputs satisfy each method's acceptance criteria.
For matched face pairs, \emph{Overall} is the fraction with matching predicted and ground-truth surface types, while \emph{Mean} is the unweighted mean of per-type accuracies.
}
\label{tab:geometry_primitive_analysis}
\begingroup
\setlength{\tabcolsep}{2.5pt}
\begin{tabular}{l|cc|cc}
\toprule
\multirow{2}{*}{Method}
& \multicolumn{2}{c|}{Chamfer Distance~$\downarrow$}
& \multicolumn{2}{c}{Primitive Accuracy~$\uparrow$} \\
& Mesh & B-rep & Overall & Mean \\
\midrule
CADDreamer & 0.0581 & 0.1096 & 0.8690 & 0.8288 \\
\textbf{UniBRep (Ours)} & \textbf{0.0319} & \textbf{0.0345} & \textbf{0.8820} & \textbf{0.8515} \\
\bottomrule
\end{tabular}
\endgroup
\end{table}

\subsection{High-Complexity and Out-of-Distribution Inputs}
\label{subsec:beyond_standard}

\begin{table*}[t]
\centering
\caption{
\textbf{Quantitative evaluation on high-complexity and out-of-distribution objects.}
\emph{DeepCAD 31+ Faces} contains 181 high-complexity models with 31 or more faces, while \emph{BrepGen OOD Furniture} contains 40 furniture objects outside the DeepCAD~\cite{Wu_2021_ICCV} training distribution.
HoLa is omitted from DeepCAD 31+ Faces because its released configuration supports at most 30 faces; CADDreamer is omitted from BrepGen OOD Furniture because it produced only one valid output (2.5\%).
The best results within each dataset are shown in \textbf{bold}.}
\label{tab:hc_ood_comparison}
\begingroup
\setlength{\tabcolsep}{4pt}
\renewcommand{\arraystretch}{1.1}
\begin{tabular}{c|l|ccc|ccc|cc|c}
\toprule
\multirow{2}{*}{Dataset}
& \multirow{2}{*}{Method}
& \multicolumn{3}{c|}{Chamfer Distance~$\downarrow$}
& \multicolumn{3}{c|}{Geometry F-score~$\uparrow$}
& \multicolumn{2}{c|}{Topology~$\uparrow$}
& \multirow{2}{*}{Valid Ratio~$\uparrow$} \\
& & Vertex & Edge & Face
& Vertex & Edge & Face
& FE & EV
& \\
\midrule

\multirow{4}{*}{\shortstack{DeepCAD~\cite{Wu_2021_ICCV}\\31+ Faces}}
& CADDreamer~\cite{li2025caddreamer}
& 0.3794 & 0.2605 & 0.1950
& 0.145 & 0.208 & 0.216
& 0.107 & 0.186
& 9.94\% \\

& Point2CAD~\cite{liu2024point2cad}
& 0.2384 & 0.1271 & 0.0751
& 0.371 & 0.475 & 0.533
& 0.362 & 0.301
& -- \\

& NVDNet~\cite{liu2024nvdnet}
& 0.2073 & 0.1123 & 0.0603
& 0.397 & 0.466 & 0.501
& 0.314 & 0.293
& -- \\

& \textbf{UniBRep (Ours)}
& \textbf{0.1408} & \textbf{0.0887} & \textbf{0.0581}
& \textbf{0.584} & \textbf{0.619} & \textbf{0.645}
& \textbf{0.534} & \textbf{0.533}
& \textbf{37.57\%} \\

\midrule

\multirow{2}{*}{\shortstack{BrepGen~\cite{xu2024brepgen}\\OOD Furniture}}
& HoLa (demo)~\cite{liu2025hola}
& 0.2365 & 0.1723 & 0.1018
& 0.354 & 0.518 & 0.544
& 0.602 & 0.350
& 22.5\% \\

& \textbf{UniBRep (Ours)}
& \textbf{0.1423} & \textbf{0.0843} & \textbf{0.0593}
& \textbf{0.704} & \textbf{0.735} & \textbf{0.742}
& \textbf{0.731} & \textbf{0.731}
& \textbf{72.5\%} \\

\bottomrule
\end{tabular}
\endgroup
\end{table*}

Table~\ref{tab:hc_ood_comparison} evaluates UniBRep beyond the standard DeepCAD benchmark along two axes: shape complexity and distribution shift.
On DeepCAD 31+ Faces, UniBRep performs best across all reported metrics, reducing face Chamfer distance from $0.1950$ to $0.0581$ and raising validity from $9.94\%$ to $37.57\%$ relative to CADDreamer.
Although validity is lower than on the standard 7--30-face benchmark, UniBRep retains a substantial advantage over the baselines on these more complex shapes.

\begin{figure*}[t]
    \centering
    \includegraphics[width=\linewidth]{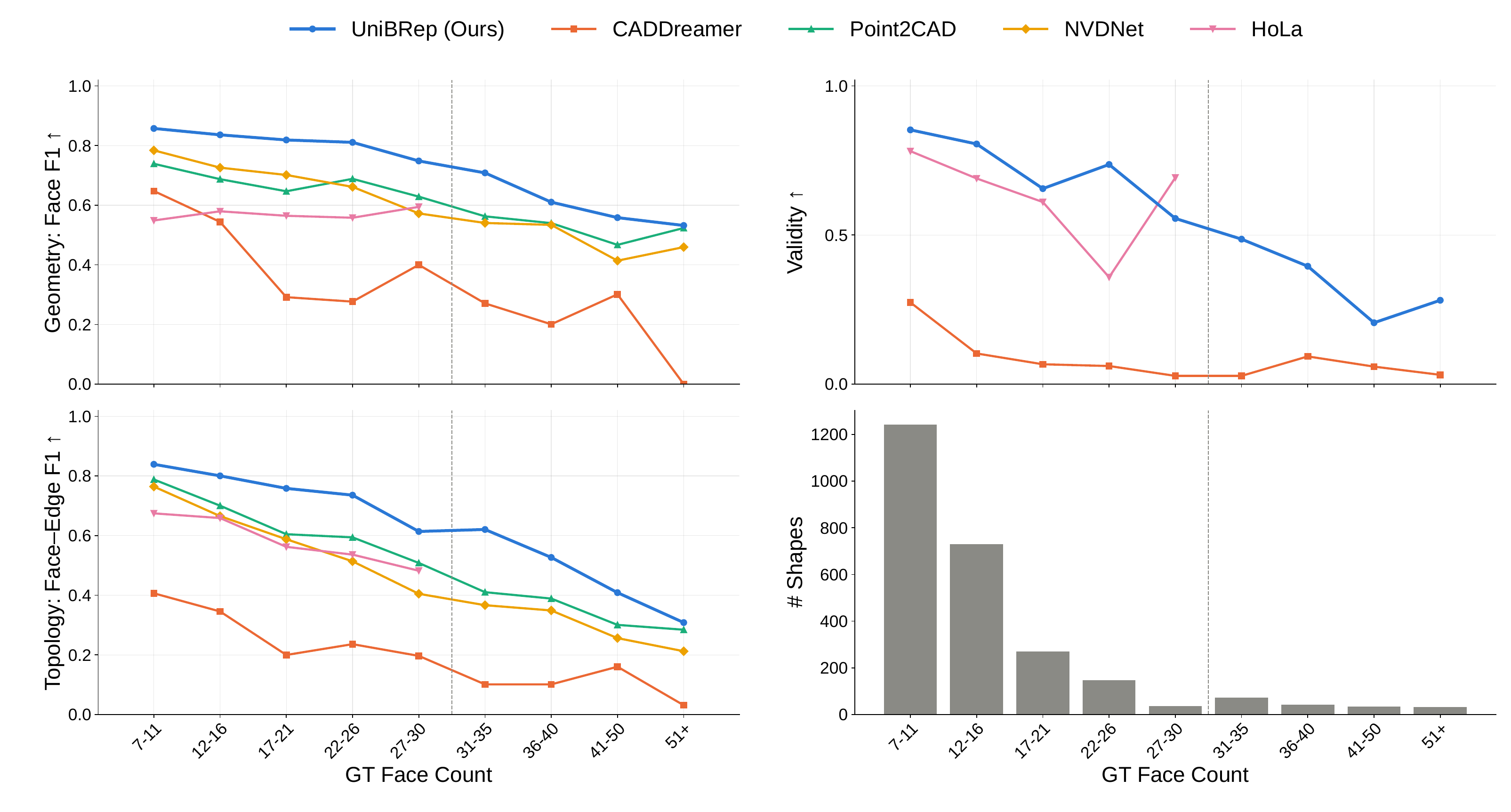}
    \vspace{-6.5mm}
    \caption{
    \textbf{Performance across B-rep face counts on DeepCAD and DeepCAD 31+ Faces~\cite{Wu_2021_ICCV}.}
    We compare UniBRep with CADDreamer, Point2CAD, NVDNet, and the HoLa demo~\cite{li2025caddreamer,liu2024point2cad,liu2024nvdnet,liu2025hola}.
    Metrics are plotted against ground-truth face count.
    Face counts are grouped into bins of five.
    For methods that generate B-reps, F-scores are averaged over valid outputs, and validity denotes the fraction of inputs yielding valid B-reps in each bin.
    Validity is omitted for Point2CAD and NVDNet because they output parametric surfaces rather than B-reps; the HoLa demo is evaluated on the matched subset of 818 test objects and supports at most 30 faces.
    }
    \label{fig:scale_plot}
\end{figure*}

\begin{figure*}[t]
    \centering
    \includegraphics[width=\linewidth]{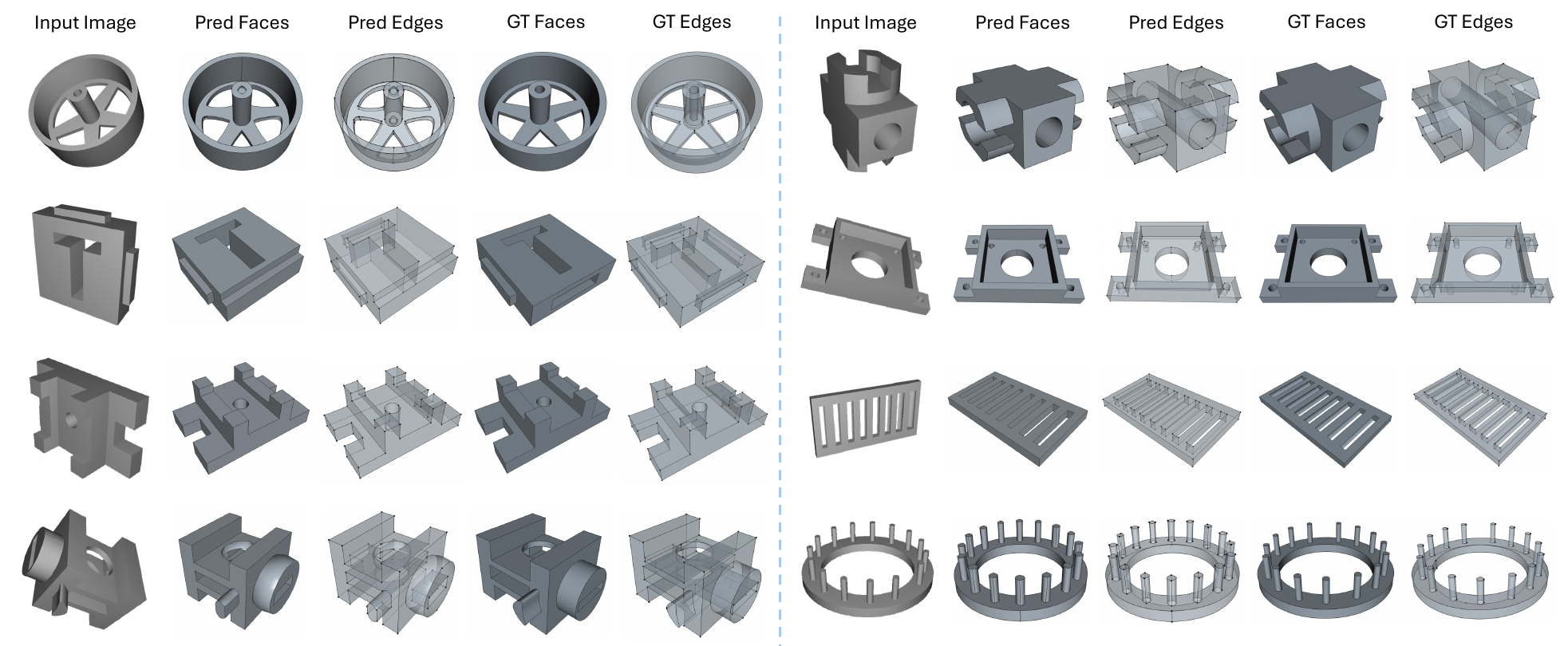}
    \vspace{-3mm}
    \caption{
    \textbf{Qualitative results on DeepCAD 31+ Faces~\cite{Wu_2021_ICCV}.}
    Each example shows the input image, UniBRep's predicted faces and edges, and the corresponding ground-truth faces and edges.
    The baselines~\cite{li2025caddreamer,liu2025hola,liu2024point2cad,liu2024nvdnet} are omitted because none produces a comparable B-rep for these examples: CADDreamer fails valid B-rep construction, the HoLa demo supports at most 30 faces, and Point2CAD and NVDNet output parametric surfaces rather than B-reps.
    }
    \label{fig:qualitative_deepcad_31plus}
\end{figure*}

\begin{figure*}[t]
    \centering
    \includegraphics[width=\linewidth]{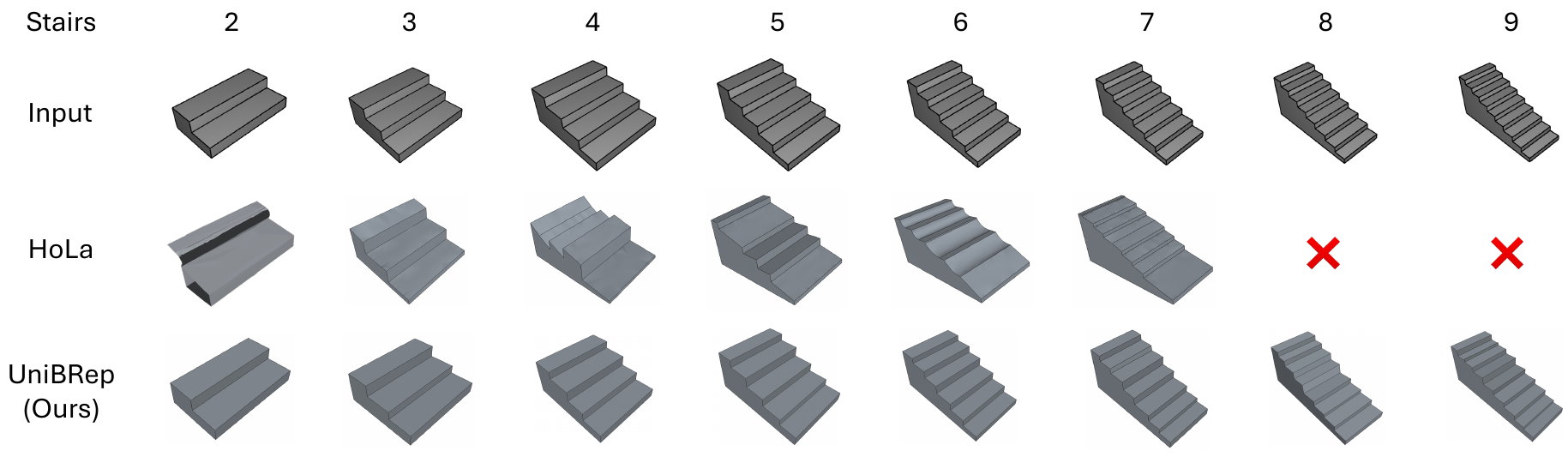}
    \vspace{-3mm}
    \caption{
    \textbf{Scalability across stair counts.}
    We vary the number of steps from two to nine.
    For the HoLa demo~\cite{liu2025hola}, the invalid two-step output is shown as a mesh, while red Xs indicate that no B-rep is produced for the eight- and nine-step inputs.
    UniBRep preserves the stepped structure and produces valid B-reps for all stair counts.
    }
    \label{fig:exp_stairs}
\end{figure*}

Fig.~\ref{fig:scale_plot} plots performance by face count.
UniBRep achieves the highest surface and face--edge topology F-scores in every bin and retains its topology advantage beyond 30 faces.
Fig.~\ref{fig:qualitative_deepcad_31plus} further shows recovery of holes, repeated thin parts, and detailed edge layouts on DeepCAD objects with 31 or more faces.
In the controlled staircase sequence of Fig.~\ref{fig:exp_stairs}, UniBRep produces valid outputs for staircases with two to nine steps, while the HoLa demo produces an invalid two-step output and no B-rep for the eight- and nine-step inputs.
These experiments demonstrate that UniBRep supports B-rep generation beyond the standard DeepCAD 30-face range without a predefined architectural face-count cap.

\begin{figure*}[t]
    \centering
    \includegraphics[width=\linewidth]{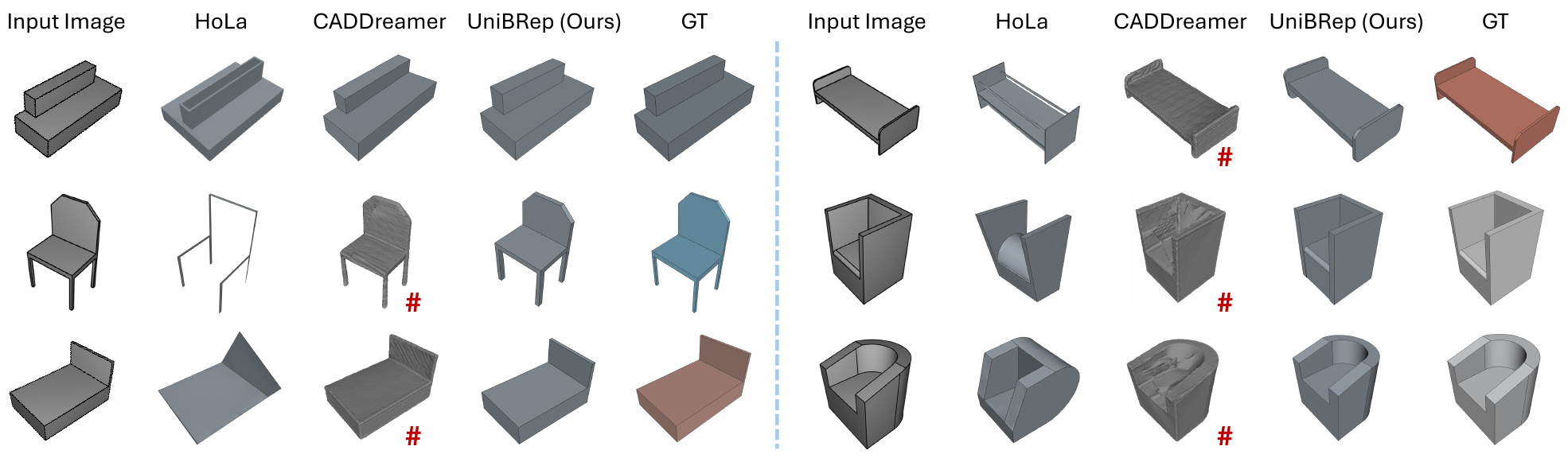}
    \vspace{-4mm}
    \caption{
    \textbf{Qualitative comparison on out-of-distribution furniture objects from BrepGen~\cite{xu2024brepgen}.}
    The HoLa demo~\cite{liu2025hola} often oversimplifies geometry or omits object parts.
    Red \# symbols mark CADDreamer~\cite{li2025caddreamer} outputs that fail valid B-rep construction.
    UniBRep better preserves overall shape, part layout, and CAD-like boundaries.
    }
    \label{fig:exp_furniture_common}
\end{figure*}

On the out-of-distribution BrepGen furniture set~\cite{xu2024brepgen}, Table~\ref{tab:hc_ood_comparison} shows that UniBRep raises validity from $22.5\%$ for the HoLa demo to $72.5\%$ and reduces face Chamfer distance from $0.1018$ to $0.0593$.
Fig.~\ref{fig:exp_furniture_common} further shows that UniBRep better preserves overall shape, part layout, and CAD-like boundaries, while the HoLa demo often simplifies or omits parts and CADDreamer frequently produces noisy meshes that fail B-rep construction.

\begin{figure}[t]
    \centering
    \includegraphics[width=\linewidth]{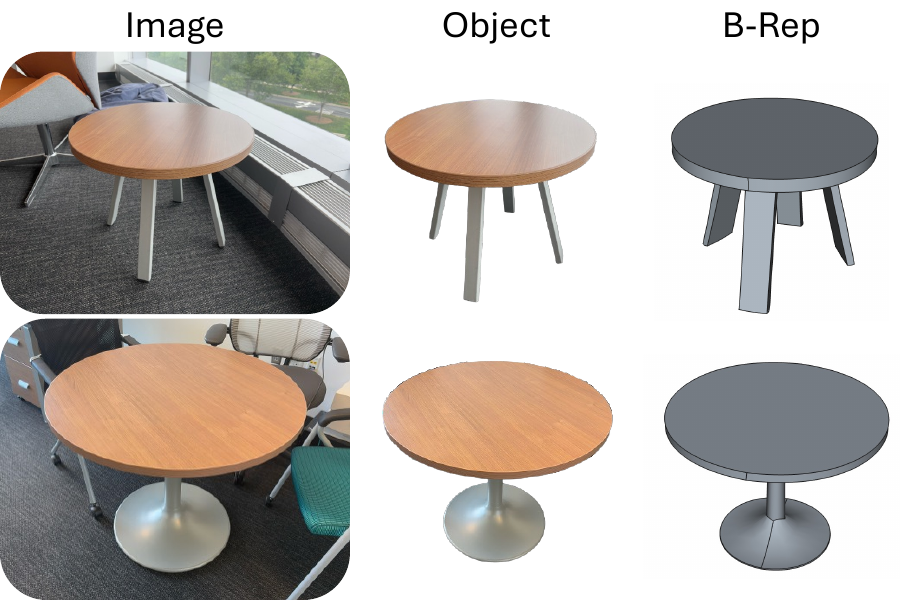}
    \caption{
    \textbf{Real-world image-conditioned B-rep generation.}
    Given foreground objects segmented from real photographs, UniBRep generates plausible geometry and constructs structured B-reps.
    }
    \label{fig:real_world_cases}
\end{figure}

Finally, Fig.~\ref{fig:real_world_cases} applies UniBRep to foreground objects segmented from real photographs.
Together, these results highlight the complementary roles of UniBRep's components: TRELLIS supplies a broad pretrained shape prior, while learned surface features guide B-rep construction from the generated geometry.

\section{Limitations and Future Work}
\label{sec:limitations}

\begin{figure*}[t]
    \centering
    \includegraphics[width=0.5\linewidth]{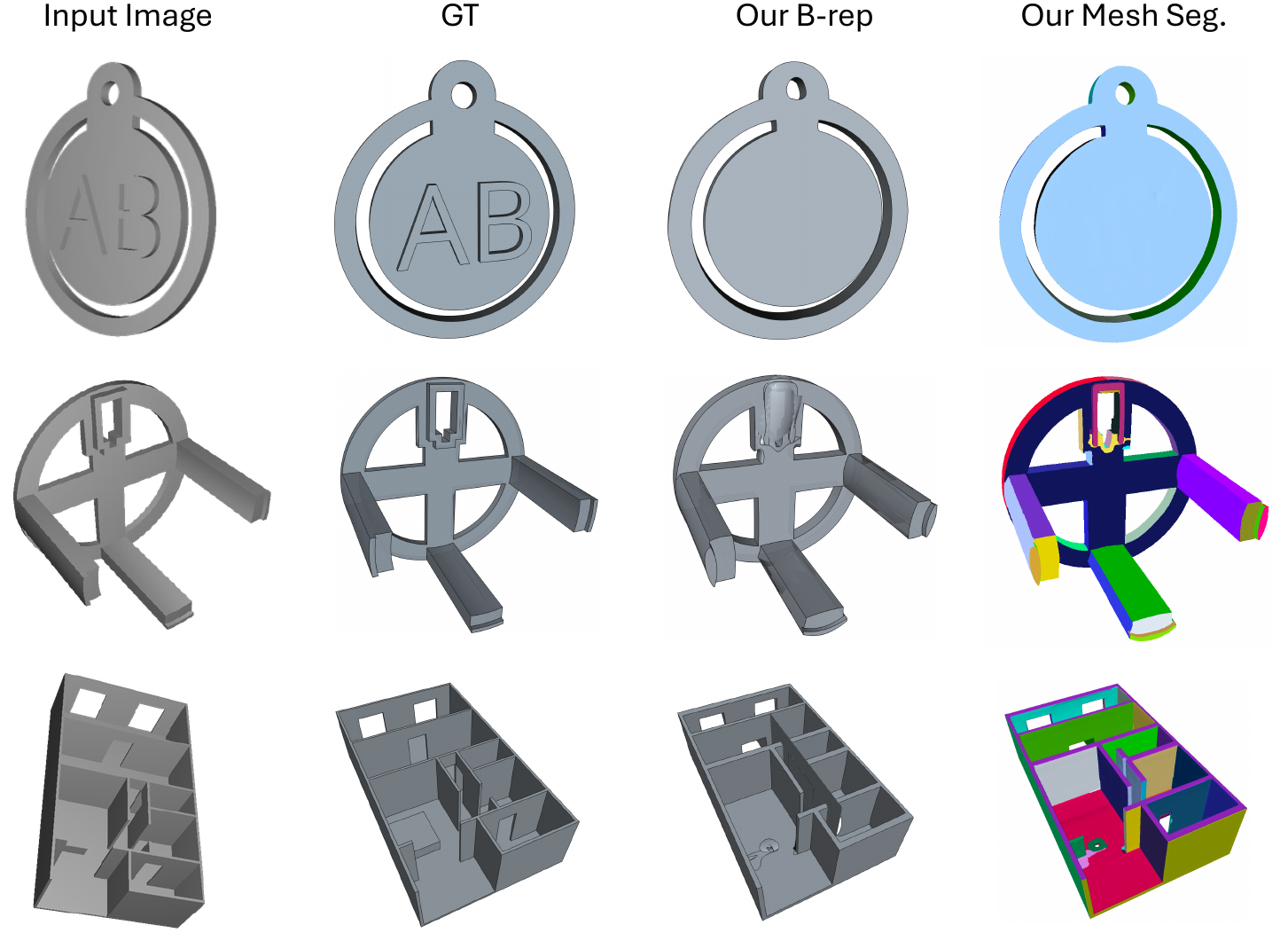}
    \caption{
    \textbf{Representative failure cases on DeepCAD 31+ Faces~\cite{Wu_2021_ICCV}.}
     From top to bottom, the examples show inaccurate geometry generation, incorrect segmentation of thin structures, and invalid B-rep assembly. CADDreamer fails to produce valid B-Reps for these cases.
    }
    \label{fig:failure_case}
\end{figure*}

Single-image B-rep generation is inherently ambiguous because occluded geometry, fine CAD details, and exact topology cannot always be inferred from one view.
Final quality depends on the generated mesh and subsequent segmentation, surface fitting, topology extraction, and CAD-kernel assembly; errors can accumulate across these stages.
On the 2,424-object DeepCAD benchmark, $0.5\%$ fail topology extraction, $4.0\%$ fail B-rep assembly, and $14.9\%$ produce outputs that fail strict validity checks; $80.5\%$ fail strict post-construction validity checks.
Thus, most failures occur after topology extraction.
Fig.~\ref{fig:failure_case} illustrates failures in geometry generation, thin-structure segmentation, and B-rep assembly.
Although UniBRep has no predefined architectural face-count limit, validity falls to $37.57\%$ on DeepCAD 31+ Faces, showing that highly complex topology remains challenging.
It nevertheless outperforms CADDreamer and the HoLa demo across all corresponding evaluations.
Quantitative evaluation remains concentrated on DeepCAD; performance on broader free-form and NURBS-heavy datasets remains untested.
The resulting B-reps support standard CAD operations but do not recover design history or parametric constraints.

Future work could incorporate richer conditioning modalities, such as multi-view images or point clouds, and expand training and evaluation to broader free-form CAD datasets.
More tightly integrating topology-aware feature learning with B-rep construction could reduce heuristic refinement and improve CAD-kernel assembly robustness.
UniBRep could also serve as an image-to-B-rep module within agentic or programmatic CAD workflows, providing explicit geometry and topology for downstream editing, program synthesis, and constraint recovery.

\section{Conclusion}
\label{sec:conclusion}

We presented UniBRep, a geometry-first framework that adapts TRELLIS to generate feature meshes combining surface geometry with spatially aligned face-separation cues.
A geometry- and feature-guided construction pipeline fits parametric surfaces, traces boundaries, and recovers explicit B-rep topology without a predefined architectural face-count limit.
Experiments demonstrate substantial improvements in geometric and topological quality and output validity over the evaluated baselines, together with generation beyond the standard 30-face range and transfer to out-of-distribution objects and real photographs.
Controlled analyses confirm the benefits of learned face features and mesh-boundary-based construction while showing that construction largely preserves the generated geometry.
Together, these results establish feature meshes as an effective bridge between broad image-to-3D priors and explicit CAD structure.

\clearpage
\clearpage
\newpage
\section*{Acknowledgements}
This research is based upon work supported by the Office of the Director of National Intelligence (ODNI), Intelligence Advanced Research Projects Activity (IARPA), via IARPA R\&D Contract No. 140D0423C0076. The views and conclusions contained herein are those of the authors and should not be interpreted as necessarily representing the official policies or endorsements, either expressed or implied, of the ODNI, IARPA, or the U.S. Government. The U.S. Government is authorized to reproduce and distribute reprints for Governmental purposes notwithstanding any copyright annotation thereon.

\bibliographystyle{eg-alpha-doi}
\bibliography{references}

\clearpage
\noWileypagenumber
\newpage

\appendix
\section{Additional Implementation Details}
\label{app:implementation_details}

\subsection{Data and Target Construction}

We construct CAD supervision targets from DeepCAD~\cite{Wu_2021_ICCV}, a filtered subset of ABC~\cite{Koch_2019_CVPR}.
Each source B-rep is normalized to \([-0.5,0.5]^3\) and tessellated with OpenCascade while preserving the correspondence between mesh triangles and their source B-rep faces.
Following TRELLIS~\cite{xiang2024structured}, we obtain sparse-structure targets by voxelizing the tessellated meshes at $64^3$ resolution.
For SLat targets, we encode multiview images into DINO features, back-project these features onto the active voxels, and pass the resulting sparse feature volume through the pretrained TRELLIS encoder.

We randomly render 150 views per model at \(512\times512\) resolution for SLat encoding and decoder supervision, along with 25 additional random views for image conditioning.
Camera viewpoints are sampled on a sphere of radius 2.0 using a Hammersley sequence with a random offset, with each camera oriented toward the object center.
Rendering uses diffuse Lambertian shading implemented in Open3D, and one conditioning view is randomly selected for each training instance.

For face-level supervision, each sampled ground-truth surface point inherits its source B-rep face identifier.
We transfer these labels to sampled predicted-mesh vertices through nearest-neighbor search and enforce a minimum number of samples per face so that small faces remain represented during feature learning.

\subsection{Architecture and Training}

The SS and SLat generators are trained with TRELLIS-style conditional flow-matching objectives at sampled timesteps.
The SLat representation has a sparse resolution of \(64^3\), with inactive voxels omitted, and the decoder progressively refines and upsamples it to predict dense geometry and feature attributes at \(256^3\) resolution.
Table~\ref{tab:module_freezing} summarizes which modules are frozen, LoRA-adapted, or trained directly.

The geometry branch follows the TRELLIS mesh-decoder training procedure with
\begin{equation}
\mathcal{L}_{\mathrm{geo}} =
0.2\mathcal{L}_{\mathrm{SSIM}} +
0.2\mathcal{L}_{\mathrm{LPIPS}} +
0.01\mathcal{L}_{\mathrm{TSDF}} +
10.0\mathcal{L}_{\mathrm{depth}},
\end{equation}
where the SSIM and LPIPS terms supervise rendered appearance, while the TSDF and depth terms supervise surface geometry.

For the contrastive feature objective defined in Sec.~\ref{subsec:feature_mesh_generation}, we follow HPNet~\cite{yan2021hpnet} and SED-Net~\cite{li2023sednet} and set
\begin{equation}
\delta_v=0.5, \qquad
\delta_d=1.5, \qquad
\lambda_{\mathrm{pull}}=\lambda_{\mathrm{push}}=1.0.
\end{equation}
For RGB supervision, following SegviGen~\cite{li2026segvigen}, we assign random colors to B-rep faces and construct ground-truth structured latents from the colored meshes for SLat-generator training and SLat-decoder input.
The decoder is supervised with an MSE loss on its predicted three-channel feature output.
Unlike the contrastive objective, RGB supervision does not explicitly encourage feature compactness within faces or separation across faces.

\begin{table}[t]
\centering
\caption{
\textbf{Module adaptation strategy.}
\emph{Frozen} modules retain pretrained TRELLIS~\cite{xiang2024structured} weights, \emph{LoRA-adapted} modules receive low-rank updates~\cite{hu2022lora}, and \emph{Trainable} modules are optimized directly.
}
\label{tab:module_freezing}
\begin{tabular}{lc}
\toprule
Component & Status \\
\midrule
SS generator input layer & Frozen \\
SS generator transformer blocks & LoRA-adapted \\
SS generator output layer & Trainable \\
SS decoder & Frozen \\
\addlinespace
SLat generator transformer blocks & LoRA-adapted \\
SLat generator output layer & Trainable \\
\addlinespace
SLat decoder geometry branch & LoRA-adapted \\
Geometry output layer & Trainable \\
SLat decoder feature branch & LoRA-adapted \\
Feature output layer & Trainable \\
\bottomrule
\end{tabular}
\end{table}

We train the SS generator on four NVIDIA A6000 GPUs for four days with batch size 4 and the SLat generator on two A6000 GPUs for three days with batch size 2.
The decoder branches are trained sequentially on two A6000 GPUs for five days with batch size 4, first optimizing the geometry branch and then freezing it before optimizing the feature branch.

\begin{table*}[t]
\centering
\small
\caption{
\textbf{Comparison with HoLa's published and public-demo results on DeepCAD~\cite{Wu_2021_ICCV}.}
The \emph{Published} row lists HoLa's reported values under the authors' original protocol.
The \emph{Matched} rows evaluate the HoLa demo and UniBRep on 818 randomly selected test objects, with a shared camera viewpoint per object and the common protocol described in Sec.~\ref{subsec:metrics}.
UniBRep receives an RGB rendering, while HoLa receives the same-view rendering with black edge overlays matching its paper.
Because the protocols are not directly comparable, the published results are included only for context.
}
\label{tab:hola_reported_vs_demo}
\resizebox{\textwidth}{!}{%
\begin{tabular}{llc|ccc|ccc|cc|c}
\toprule
\multirow{2}{*}{Method}
& \multirow{2}{*}{Protocol}
& \multirow{2}{*}{$N$}
& \multicolumn{3}{c|}{Chamfer Distance~$\downarrow$}
& \multicolumn{3}{c|}{Geometry F-score~$\uparrow$}
& \multicolumn{2}{c|}{Topology~$\uparrow$}
& \multirow{2}{*}{Valid Ratio~$\uparrow$} \\
& & & Vertex & Edge & Face & Vertex & Edge & Face & FE & EV & \\
\midrule
HoLa~\cite{liu2025hola}
& Published & 2,424
& 0.0684 & 0.0182 & 0.0101
& 0.790 & 0.820 & 0.860
& 0.810 & 0.800
& 86.37\% \\
\midrule
HoLa (demo)~\cite{liu2025hola}
& Matched & 818
& 0.3227 & 0.1737 & 0.1028
& 0.213 & 0.420 & 0.560
& 0.651 & 0.340
& 70.66\% \\

UniBRep (Ours)
& Matched & 818
& 0.1277 & 0.0593 & 0.0327
& 0.754 & 0.802 & 0.853
& 0.816 & 0.782
& 81.30\% \\
\bottomrule
\end{tabular}%
}
\end{table*}

\begin{table*}[t]
\centering
\caption{
\textbf{Sensitivity to B-rep construction parameters on DeepCAD~\cite{Wu_2021_ICCV}.}
Each parameter row varies only the indicated reconstruction setting from its default on 200 randomly selected test objects.
\emph{Geometry seg.} jointly scales the boundary-strength and normal-consistency thresholds ($g_{\mathrm{thresh}}{=}15.0$, $n_{\mathrm{thresh}}{=}0.25$).
\emph{Feature seg.} scales the mean-shift bandwidth used for feature-guided refinement (0.85).
\emph{Primitive support} changes the minimum RANSAC support-ratio threshold (0.9), while \emph{RANSAC $\epsilon$} scales the maximum mean point-to-surface residual (0.002).
\emph{Multi-seed} reports the best of five post-processing runs.
}
\label{tab:rebuttal_sensitivity}
\resizebox{\linewidth}{!}{%
\begin{tabular}{lccccccccc}
\toprule
Setting & CD(V)~$\downarrow$ & CD(E)~$\downarrow$ & CD(F)~$\downarrow$ & F-score(V)~$\uparrow$ & F-score(E)~$\uparrow$ & F-score(F)~$\uparrow$ & FE~$\uparrow$ & EV~$\uparrow$ & Valid Ratio~$\uparrow$ \\
\midrule
Default & 0.1231 & 0.0531 & 0.0291 & 0.765 & 0.807 & 0.867 & 0.826 & 0.795 & 78.5\% \\%
Geometry seg. $\times 0.75$ & 0.1243 & 0.0543 & 0.0293 & 0.760 & 0.798 & 0.853 & 0.807 & 0.777 & 74.0\% \\
Geometry seg. $\times 1.25$ & 0.1232 & 0.0553 & 0.0300 & 0.758 & 0.803 & 0.863 & 0.831 & 0.802 & 76.5\% \\
Feature seg. $\times 0.75$ & 0.1234 & 0.0536 & 0.0299 & 0.759 & 0.802 & 0.860 & 0.828 & 0.792 & 79.0\% \\
Feature seg. $\times 1.25$ & 0.1295 & 0.0592 & 0.0328 & 0.736 & 0.793 & 0.853 & 0.814 & 0.776 & 81.0\% \\
Primitive support $+ 0.02$ & 0.1244 & 0.0557 & 0.0311 & 0.761 & 0.803 & 0.856 & 0.824 & 0.795 & 78.5\% \\
Primitive support $- 0.02$ & 0.1250 & 0.0561 & 0.0307 & 0.757 & 0.805 & 0.861 & 0.826 & 0.796 & 81.0\% \\
RANSAC $\epsilon \times 0.5$ & 0.1184 & 0.0534 & 0.0290 & 0.768 & 0.813 & 0.864 & 0.827 & 0.799 & 79.5\% \\
RANSAC $\epsilon \times 2.0$ & 0.1130 & 0.0534 & 0.0291 & 0.784 & 0.821 & 0.871 & 0.842 & 0.812 & 73.5\% \\
\midrule
Multi-seed & 0.1163 & 0.0510 & 0.0279 & 0.769 & 0.818 & 0.872 & 0.834 & 0.801 & 78.5\% \\
\bottomrule
\end{tabular}
}
\end{table*}

\subsection{B-Rep Construction Details}

We normalize each generated feature mesh to \([-0.5,0.5]^3\) before construction.

\paragraph{Geometry-guided segmentation and analytic fitting.}
We partition the mesh into connected candidate face regions using region growing~\cite{adams1994regiongrowing}.
The boundary strength of each mesh edge is computed from the angle between its incident faces and robustly normalized; we use the default threshold \(g_{\mathrm{thresh}}=15.0\).
During region expansion, normal consistency is measured as
\begin{equation}
d_n(\mathbf{n}_{r},\mathbf{n}_{f}) = 1-\mathbf{n}_{r}^{\mathsf{T}}\mathbf{n}_{f},
\end{equation}
where \(\mathbf{n}_{r}\) and \(\mathbf{n}_{f}\) are the current region normal and candidate-face normal.
A candidate mesh face is added when \(d_n<n_{\mathrm{thresh}}\), with \(n_{\mathrm{thresh}}=0.25\).

For each candidate region, we use CGAL RANSAC~\cite{schnabel2007ransac} to fit planes, cylinders, cones, spheres, and tori in sequence.
We perform ten trials per primitive type and accept the first type whose support ratio exceeds \(0.9\) and whose mean point-to-surface residual is below \(0.002\).
If none passes, we accept a planar fallback when its support ratio exceeds \(0.85\) and its mean residual is below \(0.005\).
Unresolved regions are passed to feature-guided refinement.
Each accepted analytic surface is parameterized and sampled on a \(32\times32\) grid in its parametric domain.

\paragraph{Feature-guided refinement and free-form fitting.}
For each region that fails initial primitive fitting, we apply mean-shift clustering to its eight-dimensional vertex features using a bandwidth of \(0.85\).
Because a cluster may contain spatially disconnected patches, we divide it into connected components according to mesh adjacency and repeat primitive fitting for each component using the same type order, trials, acceptance criteria, and planar fallback.
A region that remains unresolved is represented by a B-spline surface.
We use FlattenAnything~\cite{zhang2024flatten} to parameterize each mesh patch in two dimensions and optimize the parameterization for 1,000 iterations.
We then sample a \(32\times32\) grid, map the samples back to 3D, and fit a B-spline surface to the resulting points.

\paragraph{Boundary recovery and B-rep assembly.}
After assigning a fitted surface to each face region, we trace shared mesh boundaries between adjacent regions to obtain candidate edge polylines without explicitly intersecting the fitted surfaces.
Following BrepGen~\cite{xu2024brepgen}, we approximate each polyline with a cubic B-spline curve and uniformly resample it at 32 locations along its arc length.
Endpoints shared by multiple curves are fixed to common vertex positions.
The resulting incidences among face regions, boundary curves, and shared endpoints determine the face--edge and edge--vertex connectivity.

For each incident face, we project the recovered 3D boundary curve onto its fitted surface and construct a p-curve using \texttt{ShapeFix\_Edge::FixAddPCurve}.
These p-curves define the trimming loops on the fitted surfaces.
We assemble and sew the trimmed faces, shared curves, vertices, and recovered incidence relations using \texttt{BRepBuilderAPI\_Sewing}.
Loop consistency, face and wire orientation, and small geometric discrepancies are handled through tolerance-based healing with \texttt{ShapeFix\_Wire} and \texttt{ShapeFix\_Face}, using a precision of \(0.01\) and a maximum tolerance of \(0.1\).

\section{Additional Evaluation Details}
\label{app:evaluation_details}

\subsection{B-Rep Validation Protocol}

Following BrepDiff~\cite{lee2025brepdiff}, we classify an assembled output as valid only if it passes all CAD-kernel and tessellation checks.
The STEP file must be readable, contain a single top-level \texttt{TopAbs\_SOLID}, and be reported as valid by \texttt{BRepCheck\_Analyzer}.
The output must also pass the wire- and edge-consistency checks adopted from BrepDiff and tessellate into a nonempty STL mesh within a 20-second timeout.
Outputs that fail any check or stall during tessellation are treated as invalid.

\begin{figure*}[t]
    \centering
    \includegraphics[width=\linewidth]{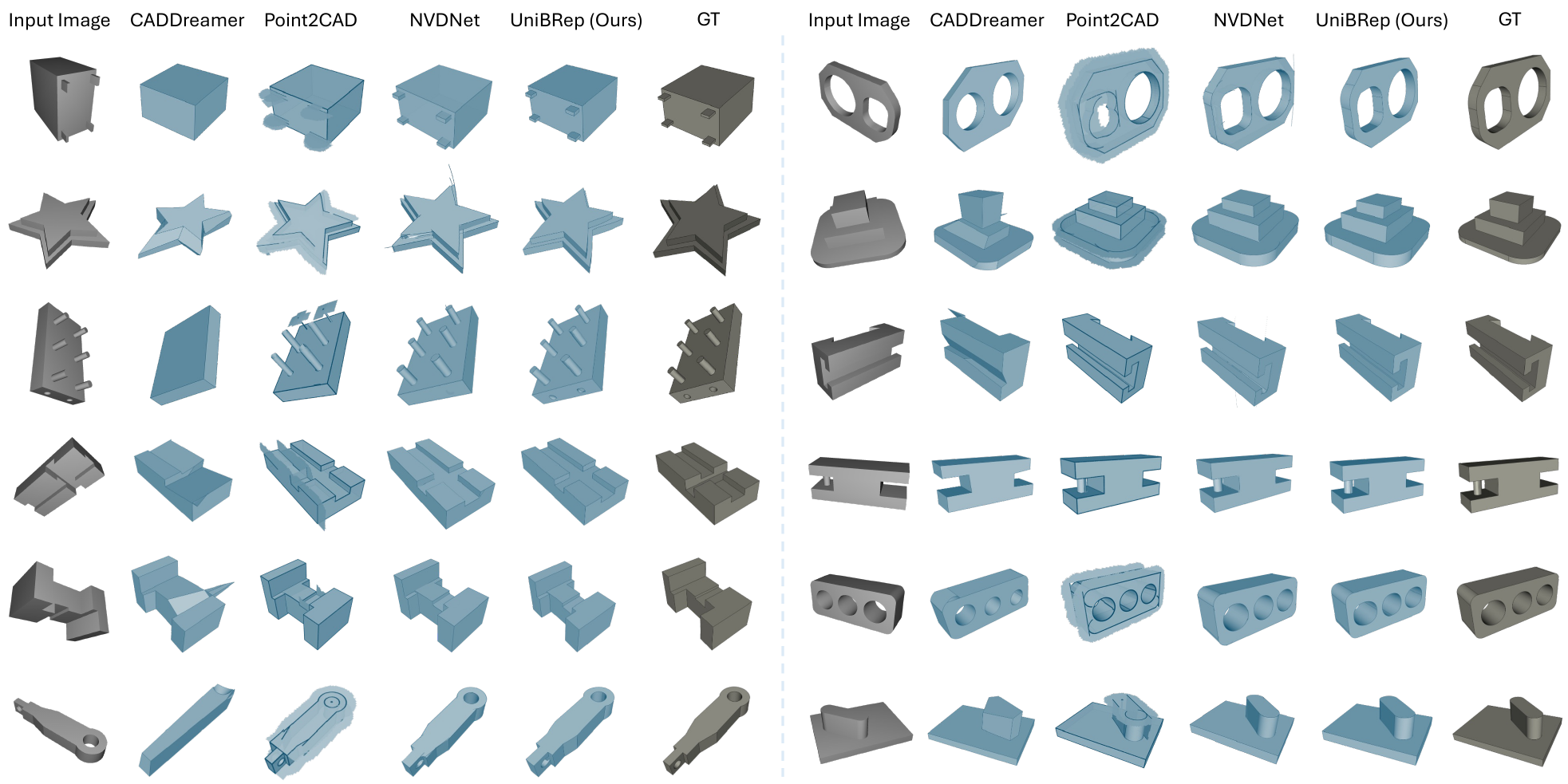}
    \caption{
    \textbf{Additional qualitative comparison on DeepCAD~\cite{Wu_2021_ICCV}.}
    We compare UniBRep with the baselines~\cite{li2025caddreamer,liu2025hola,liu2024point2cad,liu2024nvdnet}.
    Outputs and ground truth are rendered from a common evaluation view that differs from the input view.
    }
    \label{fig:qualitative_deepcad}
\end{figure*}

\subsection{HoLa Published Results and Public-Demo Evaluation}

Table~\ref{tab:hola_reported_vs_demo} presents HoLa's published results~\cite{liu2025hola} alongside our matched evaluation of its public demo and UniBRep.
We randomly select 818 objects from the standard 7--30-face DeepCAD benchmark and pair each with one conditioning viewpoint sampled using the spherical camera setup in Appendix~\ref{app:implementation_details}.
For each object--viewpoint pair, UniBRep receives an RGB rendering, while HoLa receives an RGB rendering from the same viewpoint with black edge overlays matching the input style shown in its paper.
Outputs from both systems are evaluated using the common protocol described in Sec.~\ref{subsec:metrics}, and all selected objects are retained regardless of output quality.

Beyond the dataset split and input appearance described above, HoLa does not release the model weights, inference and evaluation code, camera-selection procedure, or preprocessing instructions needed for exact reproduction.
The public demo performs substantially below HoLa's published results, but we cannot isolate whether the discrepancy arises from camera settings, rendering, preprocessing, or evaluation.
We therefore use the matched demo comparison in the main paper and include the published results only for context.

\section{Reconstruction Parameter Sensitivity}
\label{app:reconstruction_sensitivity}

Table~\ref{tab:rebuttal_sensitivity} evaluates sensitivity to the principal reconstruction parameters on 200 randomly selected DeepCAD test objects.
Geometry and topology metrics remain stable under moderate perturbations, with valid ratios ranging from $73.5\%$ to $81.0\%$.
Repeating post-processing five times improves the best-run face Chamfer distance only from \(0.0291\) to \(0.0279\) and does not increase the valid count over the default run.
We therefore use a single post-processing run per input for all main results.

\section{Additional Qualitative Results}
\label{app:additional_qualitative_results}

Fig.~\ref{fig:qualitative_deepcad} presents additional qualitative comparisons on complex DeepCAD objects.
CADDreamer~\cite{li2025caddreamer} often captures the coarse layout but distorts surfaces, misses thin structures, or introduces spurious parts.
UniBRep more faithfully recovers holes, protrusions, block structures, and sharp CAD boundaries, producing B-reps that more closely match the ground truth.

\end{document}